\documentclass{article} 
\usepackage{iclr2027_conference,times}

\usepackage{amsmath,amsfonts,bm}

\def\eqref#1{equation~\ref{#1}}

\def\1{\bm{1}}

\DeclareMathAlphabet{\mathsfit}{\encodingdefault}{\sfdefault}{m}{sl}
\SetMathAlphabet{\mathsfit}{bold}{\encodingdefault}{\sfdefault}{bx}{n}

\usepackage{hyperref}
\usepackage{url}
\usepackage{enumitem}
\usepackage{graphicx}
\usepackage{algorithm}
\usepackage{algpseudocode}
\usepackage{amsmath}
\usepackage{booktabs}
\usepackage{multirow}
\usepackage[table]{xcolor}
\usepackage{makecell}
\usepackage{array}
\usepackage{tabularx}
\usepackage{subcaption}
\usepackage{placeins}
\usepackage{needspace}
\newcommand{\AblUpRed}[1]{\textcolor{red}{\scriptsize($\uparrow$#1)}}
\newcommand{\AblDownRed}[1]{\textcolor{red}{\scriptsize($\downarrow$#1)}}
\newcommand{\AblUpBlue}[1]{\textcolor{blue}{\scriptsize($\uparrow$#1)}}
\newcommand{\AblDownBlue}[1]{\textcolor{blue}{\scriptsize($\downarrow$#1)}}
\newcommand{\AblEq}[1]{\textcolor{black}{\scriptsize(=#1)}}
\newcommand{\AblCell}[2]{\makecell[c]{#1\\#2}}

\title{Cache-Aware Joint Router Adaptation for Memory-Efficient MoE Inference}

\author{%
\textbf{Zhenhe Wu$^{1,2}$\footnotemark[2], Yaping Jin$^{1}$\footnotemark[2], Qinghua Xing$^{1}$, Hang Zhou$^{3}$, Wei He$^{4}$,} \\
\textbf{Xianjie Wu$^{5}$, Xianfu Cheng$^{2}$, Jian Yang$^{2}$, Hanting Chen$^{1*}$} \\
$^{1}$Huawei Technologies, $^{2}$Beihang University, $^{3}$Tianjin University, \\
$^{4}$University of Sydney, $^{5}$Beijing Information Science \& Technology University\\
\texttt{\{wuzhenhe2,chenhanting\}@huawei.com}
}

\iclrfinalcopy 
\begin{document}

\maketitle
\lhead{}
\renewcommand{\headrulewidth}{0pt}
\ificlrfinal
\renewcommand{\thefootnote}{\fnsymbol{footnote}}
\footnotetext{$^*$Corresponding author.}
\footnotetext{$^\dagger$These authors contributed equally to this work.}
\fi

\begin{abstract}
Mixture-of-Experts (MoE) models activate few experts per token, yet their full expert sets can exceed GPU memory and require repeated weight transfers during decoding. We formulate expert-cache management as a model-side algorithmic problem and propose cache-aware post-training that jointly adapts the MoE backbone and lightweight auxiliary routers while preserving the native inference-time Top-\(K\) rule. The update-only \textbf{Temporal Router} learns same-layer retention across tokens without proactive loading. The full \textbf{Spatio-Temporal Router} adds a \textbf{Spatio Router} that uses the causal predecessor's hidden state to refine the temporal cache before target-layer access. We evaluate both modes on Qwen3 and GPT-OSS across GSM8K, MATH, and CommonsenseQA. Temporal Router consistently improves hit rate and reduces expert-weight traffic over matched LM-only baselines. On Qwen3, the full mode improves adjusted hit rate by 1.15--18.03 points and reduces traffic by 4.6--53.3\% relative to the strongest evaluated prefetching baseline; GPT-OSS results are competitive but task-dependent. Auxiliary-only training preserves baseline accuracy but yields modest coverage gains; joint post-training achieves substantially higher coverage. Sensitivity analyses distinguish the effects of cache capacity, refinement budget, and cache-loss weight on coverage, traffic, and quality.
\end{abstract}

\section{Introduction}
Mixture-of-Experts (MoE) models increase capacity by activating only a small expert subset per token. This computational sparsity does not eliminate weight-storage costs: when experts exceed an inference worker's GPU memory, decoding requires repeated transfers from slower memory. Cache management must decide which experts to retain and when to fetch them.

Existing approaches exploit expert-access locality through caching and prefetching. Classical LRU, LFU, and LRFU use recency or frequency, while MoE-specific methods exploit access traces, hidden-state predictors, retrieved expert maps, or persistent expert sets. Recent STEP and ST-MoE systems combine temporal and cross-layer signals for prefetching~\citep{liu2026step,zhao2026stmoe}. This motivates a different question: can post-training jointly adapt expert demand and cache priorities, rather than only predict or schedule an existing access pattern?

We address this question with a cache-aware objective that jointly adapts the MoE backbone and lightweight auxiliary routers. The \textbf{Temporal Router} scores resident experts for retention after each layer access, anticipating reuse at the same layer by later tokens. The optional \textbf{Spatio Router} uses the causal predecessor's hidden state to refine the cache before target-layer access. Native routing still selects the executed experts; auxiliary routing controls residency. The update-only Temporal Router incurs no proactive traffic, while the full \textbf{Spatio-Temporal Router} stacks bounded pre-access refinement on the same retention stage.

Across two MoE backbones and three reasoning tasks, Temporal Router improves hit rate and traffic over replacement policies on matched LM-only post-trained backbones. The full mode leads evaluated prefetchers on all Qwen3 tasks in adjusted hit rate and traffic, with task-dependent GPT-OSS results. Ablations support joint adaptation and temporal retention; budget sweeps reveal the coverage--traffic trade-off. Our contributions are:
\begin{itemize}[leftmargin=1.5em, itemsep=0.4em, topsep=0.2em]
\item A model-side cache-aware post-training formulation that jointly adapts native routing behavior and auxiliary cache priorities without changing the inference-time Top-\(K\) rule.
\item A stackable two-stage algorithm with independently deployable Temporal Router retention and optional causal Spatio Router refinement.
\item Evaluation on two backbones and three benchmarks, including adaptation, component, capacity, refinement-budget, and loss-weight ablations under proactive-load-aware accounting.
\end{itemize}

\section{Problem Formulation}
We consider autoregressive decoding with \(L\) MoE layers and \(N\) routed experts per layer. At token \(t\) and layer \(l\), the native router maps hidden state \(h_{t,l}\) to
\[
p^R_{t,l}=\operatorname{softmax}(W^R_lh_{t,l})\in\mathbb{R}^{N},
\qquad S_{t,l}=\operatorname{TopK}(p^R_{t,l}).
\]
Each layer maintains a cache \(\mathcal{C}_{t,l}\subseteq\{1,\ldots,N\}\) with \(|\mathcal{C}_{t,l}|=B\) after initialization, where \(B\geq K\). Here, \(\mathcal{C}_{t,l}\) denotes the resident experts before access. Every selected expert in \(S_{t,l}\setminus\mathcal{C}_{t,l}\) must be demand-loaded before its computation can proceed. Auxiliary routers do not override \(S_{t,l}\). Preserving native routing means retaining the Top-\(K\) rule, not freezing its parameters.

Cache decisions occur at two points. \emph{Post-access retention} keeps experts for later tokens and cannot prevent misses at the completed access. \emph{Pre-access refinement} uses causal information to improve residency before a target access. Temporal Router uses retention only; Spatio-Temporal Router adds refinement before the same retention stage. We optimize the resulting transfer demand while tracking task quality, independently of a particular runtime scheduler.

\section{Method}
\subsection{Unified Cache-Routing Framework}
\label{sec:unified_framework}
\begin{figure*}[!tbp]
\centering
\includegraphics[width=\textwidth]{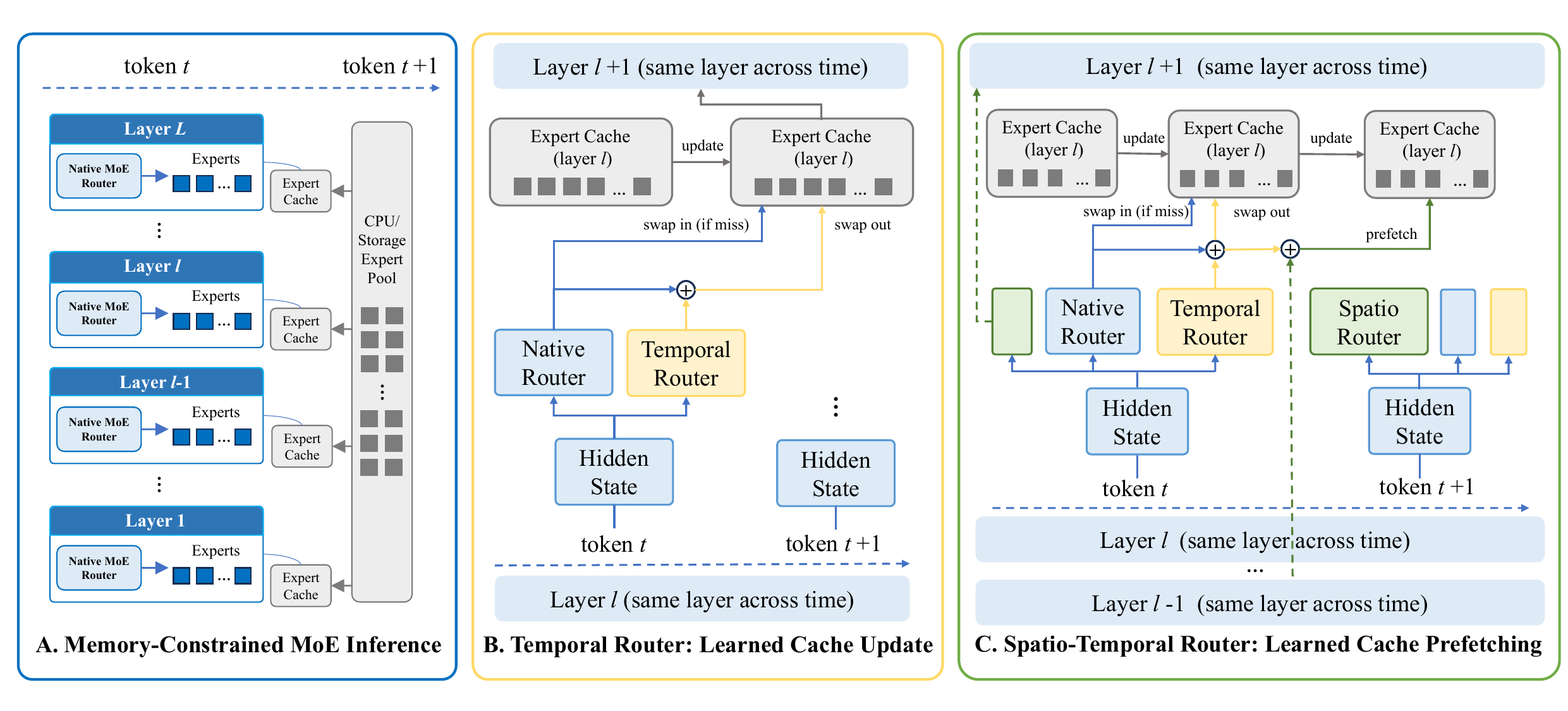}
\caption{Overview of cache-aware MoE inference. (A) Each layer maintains a bounded GPU cache backed by a host/storage expert pool. (B) Temporal Router retains experts after access for later tokens at the same layer. (C) Spatio-Temporal Router adds causal pre-access refinement, followed by the same temporal update.}
\label{fig:overview}
\end{figure*}

Figure~\ref{fig:overview} shows the cache lifecycle; Appendix~\ref{app:architecture} compares the architectures. The update-only mode accesses the temporal carry-over cache \(\mathcal{C}^{T}_{t,l}\) directly; the full mode refines it into \(\widetilde{\mathcal{C}}^{ST}_{t,l}\). After native expert computation, both apply the same retention rule to obtain \(\mathcal{C}^{T}_{t+1,l}\).

We combine normalized router distributions by equal-weight addition, without a learned fusion module. For priority \(s\in\mathbb{R}^{N}\), let \(\theta_B(s)\) be its \(B\)-th largest entry. A soft Top-\(B\) membership surrogate and the shared coverage loss are
\[
m_e(s)=\sigma\!\left(\frac{s_e-\theta_B(s)}{\tau}\right),
\qquad
\ell_{\mathrm{cache}}(s,q)=\sum_{e=1}^{N}q_e\bigl(1-m_e(s)\bigr).
\]
Using the full native-router target distribution \(q\), rather than hard Top-\(K\) labels, retains uncertainty near the selection boundary and supplies dense supervision. This soft objective learns relative priorities, not an unconstrained physical cache: the inference rules below determine which experts may enter and enforce the exact capacity. The two modes differ in their causal inputs, supervision targets, and when the resulting priority is applied. Retention is specified in Algorithm~\ref{alg:update-cache}; full inference appears in Appendix~\ref{app:algorithms}.

\subsection{Temporal Router: Post-Access Retention}
\label{sec:temporal_router}
Temporal Router predicts same-layer demand at the next token and combines it with current native routing:
\[
p^T_{t,l}=\operatorname{softmax}(W^T_lh_{t,l}),
\qquad s^T_{t,l}=p^R_{t,l}+p^T_{t,l}.
\]
The next-token native distribution supervises same-layer coverage through
\[
\ell^T_{t,l}=\sum_{e=1}^{N}p^R_{t+1,l,e}\bigl(1-m_e(s^T_{t,l})\bigr).
\]
With \(\Omega_T\) containing positions with a valid next-token target, training minimizes
\[
\mathcal{L}_{T}=\frac{1}{|\Omega_T|}\sum_{(t,l)\in\Omega_T}
\ell^T_{t,l},
\qquad
\mathcal{L}=\mathcal{L}_{LM}+\lambda_T\mathcal{L}_{T}.
\]
At inference, the native router accesses \(\mathcal{C}^{T}_{t,l}\), demand-loads missing selected experts, and executes the layer. The post-access operator then forms
\[
\mathcal{C}^{T}_{t+1,l}=\mathcal{U}_{B}\!\left(\mathcal{C}^{T}_{t,l},S_{t,l},s^T_{t,l}\right).
\]
The operator inserts selected experts that were demand-loaded for the current computation, protects all experts in \(S_{t,l}\), and evicts the lowest-priority non-selected residents when necessary. It does not admit arbitrary high-scoring experts from the external pool: every insertion is already available from the completed access. Temporal Router therefore learns retention without introducing proactive transfers. Algorithm~\ref{alg:update-cache} specifies this shared update.

\begin{algorithm}[!t]
\caption{Shared Post-Access Cache Update \(\mathcal{U}_{B}\)}
\label{alg:update-cache}
\begin{algorithmic}[1]
\Function{UpdateCache}{$\mathcal{C},S,s,B$}
\State $\mathcal{C}^{+}\gets\mathcal{C}$
\State $\mathcal{D}\gets S\setminus\mathcal{C}^{+}$ \Comment{demand-loaded selected experts}
\ForAll{$e\in\mathcal{D}$}
    \If{$|\mathcal{C}^{+}|<B$}
        \State $\mathcal{C}^{+}\gets\mathcal{C}^{+}\cup\{e\}$
    \Else
        \State $\mathcal{V}\gets\mathcal{C}^{+}\setminus S$ \Comment{protect current selected experts}
        \State $v\gets\arg\min_{u\in\mathcal{V}}s_u$
        \State $\mathcal{C}^{+}\gets(\mathcal{C}^{+}\setminus\{v\})\cup\{e\}$
    \EndIf
\EndFor
\State \Return $\mathcal{C}^{+}$
\EndFunction
\end{algorithmic}
\end{algorithm}

\subsection{Spatio-Temporal Router: Pre-Access Refinement}
\label{sec:spatio_temporal_router}
The full mode adds a Spatio Router at source layer \(l\), producing \(p^S_{t,l}=\operatorname{softmax}(W^S_lh_{t,l})\) for the next MoE position in causal execution order. For target \((t,l)\), its predecessor is
\[
\rho(t,l)=
\begin{cases}
(t-1,L), & l=1,\\
(t,l-1), & l>1.
\end{cases}
\]
Thus, the first layer uses the preceding token's final-layer prediction; other layers use the preceding layer of the same token. The pre-access priority combines temporal carry-over with this prediction:
\[
a^T_{t,l}=p^R_{t-1,l}+p^T_{t-1,l},
\qquad
s^{ST}_{t,l}=a^T_{t,l}+p^S_{\rho(t,l)}.
\]
Equivalently, the first-layer wrap-around and within-token cases are
\[
s^{ST}_{t,l}=
\begin{cases}
p^R_{t-1,1}+p^T_{t-1,1}+p^S_{t-1,L}, & l=1,\\
p^R_{t-1,l}+p^T_{t-1,l}+p^S_{t,l-1}, & l>1.
\end{cases}
\]
Each sum refers to experts at the target layer; the Spatio Router is indexed by the source that produces its prediction. Because this priority is available before target access, its training target is the current native distribution:
\[
\ell^{ST}_{t,l}=\sum_{e=1}^{N}p^R_{t,l,e}\bigl(1-m_e(s^{ST}_{t,l})\bigr).
\]
Averaging over valid targets \(\Omega_{ST}\) gives
\[
\mathcal{L}_{ST}=\frac{1}{|\Omega_{ST}|}\sum_{(t,l)\in\Omega_{ST}}
\ell^{ST}_{t,l},
\qquad
\mathcal{L}=\mathcal{L}_{LM}+\lambda_{ST}\mathcal{L}_{ST}.
\]
Here, \(\Omega_{ST}\) requires both a previous-token same-layer state and a causal predecessor.

Before access, the full mode examines the top-\(R\) candidates under \(s^{ST}_{t,l}\), in descending priority. A non-resident candidate replaces the lowest-priority resident only if its score is higher; each insertion is a proactive load. The native router then selects \(S_{t,l}\), demand-loads remaining misses, and executes the layer. The shared temporal update produces \(\mathcal{C}^{T}_{t+1,l}\) (Algorithm~\ref{alg:spatio-temporal-router}, Appendix~\ref{app:algorithms}). The soft Top-\(B\) objective learns priorities, whereas the inference-only Top-\(R\) filter limits candidates rather than actual loads; \(R\) can be changed without retraining.

\subsection{Training and Deployment Modes}
\label{sec:training_deployment}
Main runs jointly train the full backbone and auxiliary routers. Native distributions in the cache loss remain differentiable, so the objective can adapt expert demand as well as cache priorities. The full mode uses \(\mathcal{L}_{ST}\), with neither a separate Temporal Router checkpoint nor an additional \(\mathcal{L}_{T}\) term. Future/target distributions provide teacher-forced training supervision only; inference uses causal hidden states and stored outputs. The LM-only reference omits auxiliary routers and optimizes \(\mathcal{L}_{LM}\) alone (\(sw=0\)). The auxiliary-only control freezes the LM-only backbone and trains only the added routers. This separates auxiliary prediction on fixed model representations from joint adaptation; it does not isolate the contribution of any one backbone component.

\paragraph{Router initialization.}
The auxiliary routers are initialized by copying native-router weights, not by random initialization:
\[
W^T_l\leftarrow W^R_l,\qquad
W^S_l\leftarrow
\begin{cases}
W^R_{l+1}, & l<L,\\
W^R_1, & l=L.
\end{cases}
\]
Temporal Router starts from the same-layer router, while Spatio Router starts from its prediction target's router, including the final-to-first-layer wrap-around. Appendix~\ref{app:implementation} details evaluation controls.

\section{Experiments}
\label{sec:experiments}
\subsection{Experimental Setup}
\paragraph{Backbone models.}
We use Qwen3-30B-A3B-Instruct-2507~\citep{qwen3technicalreport} and GPT-OSS-20B~\citep{agarwal2025gpt}. Qwen3 has 48 MoE layers, approximately 30.5B total and 3.3B activated parameters, and 128 routed experts per layer with top-8 selection. GPT-OSS has 24 layers, approximately 21B total and 3.6B activated parameters, and 32 experts per layer with top-4 selection. Both use the same post-training and evaluation pipeline.

\paragraph{Datasets.}
GSM8K contains 7,473 training and 1,319 test grade-school mathematics problems~\citep{cobbe2021training}. MATH contains 7,500 training and 5,000 test competition-level problems~\citep{hendrycks2021measuring}. CommonsenseQA contains 9,741 training, 1,221 validation, and 1,140 hidden-test questions~\citep{talmor2019commonsenseqa}; we evaluate on validation because test labels are not public. Accuracy uses exact match for GSM8K and CommonsenseQA and symbolic or numeric matching for MATH.

\begin{table*}[t]
\centering
\setlength{\tabcolsep}{3.8pt}
\resizebox{\textwidth}{!}{
\begin{tabular}{l|l|l|c|cccc|cccc|cccc}
\toprule
\multirow{2}{*}{\textbf{Method}} &
\multirow{2}{*}{\makecell{\textbf{Decision}\\\textbf{Availability}}} &
\multirow{2}{*}{\textbf{Model}} &
\multirow{2}{*}{\makecell{\textbf{Added}\\\textbf{Params.}}} &
\multicolumn{4}{c|}{\textbf{GSM8K}} &
\multicolumn{4}{c|}{\textbf{MATH}} &
\multicolumn{4}{c}{\textbf{CommonsenseQA}} \\
\cmidrule(lr){5-8}
\cmidrule(lr){9-12}
\cmidrule(lr){13-16}
& & & &
Acc. & Hit & Adj. Hit & Load &
Acc. & Hit & Adj. Hit & Load &
Acc. & Hit & Adj. Hit & Load \\
\midrule

\rowcolor{gray!15}
\multicolumn{16}{c}{\textbf{Cache-Update Methods}} \\
\midrule

\multirow{2}{*}{MoE / LRU}
& \multirow{2}{*}{Prev. token} & Qwen3 & --
& \textbf{85.44} & 61.19 & -- & 1407
& \textbf{58.22} & \underline{64.30} & -- & \underline{1294}
& \textbf{87.39} & 56.44 & -- & 1579 \\
\cmidrule(lr){3-16}
& & GPT-OSS & --
& \underline{61.87} & 65.58 & -- & 1645
& \textbf{43.74} & 67.15 & -- & 1569
& \underline{84.68} & 65.10 & -- & 1668 \\
\cmidrule(lr){1-16}

\multirow{2}{*}{MoE / LFU}
& \multirow{2}{*}{Prev. token} & Qwen3 & --
& \textbf{85.44} & 59.99 & -- & 1450
& \textbf{58.22} & 63.00 & -- & 1341
& \textbf{87.39} & \underline{58.27} & -- & \underline{1512} \\
\cmidrule(lr){3-16}
& & GPT-OSS & --
& \underline{61.87} & 68.63 & -- & 1499
& \textbf{43.74} & 65.37 & -- & 1654
& \underline{84.68} & \underline{69.29} & -- & \underline{1467} \\
\cmidrule(lr){1-16}

\multirow{2}{*}{MoE / LRFU}
& \multirow{2}{*}{Prev. token} & Qwen3 & --
& \textbf{85.44} & \underline{62.67} & -- & \underline{1353}
& \textbf{58.22} & 64.29 & -- & \underline{1294}
& \textbf{87.39} & 57.89 & -- & 1526 \\
\cmidrule(lr){3-16}
& & GPT-OSS & --
& \underline{61.87} & \underline{69.55} & -- & \underline{1455}
& \textbf{43.74} & \underline{68.66} & -- & \underline{1497}
& \underline{84.68} & 68.34 & -- & 1513 \\
\cmidrule(lr){1-16}

\multirow{2}{*}{\textbf{\makecell{Temporal\\Router}}}
& \multirow{2}{*}{Prev. token} & Qwen3 & 12.6M
& \textbf{85.44} & \textbf{73.13} & -- & \textbf{974}
& \underline{57.82} & \textbf{75.00} & -- & \textbf{906}
& \underline{86.16} & \textbf{91.61} & -- & \textbf{304} \\
\cmidrule(lr){3-16}
& & GPT-OSS & 2.2M
& \textbf{63.99} & \textbf{72.84} & -- & \textbf{1298}
& \textbf{43.74} & \textbf{71.84} & -- & \textbf{1345}
& \textbf{85.83} & \textbf{71.54} & -- & \textbf{1360} \\

\midrule
\rowcolor{gray!15}
\multicolumn{16}{c}{\textbf{Prefetching Methods}} \\
\midrule

\multirow{2}{*}{\makecell{Least-Stale\\(SpecMD)}}
& \multirow{2}{*}{Prev. token} & Qwen3 & --
& \textbf{85.44} & 68.52 & 31.12 & 5496
& \textbf{58.22} & 70.02 & 31.55 & 5505
& \textbf{87.39} & 67.56 & 30.17 & 5667 \\
\cmidrule(lr){3-16}
& & GPT-OSS & --
& \underline{61.87} & 70.07 & 35.59 & 6058
& \textbf{43.74} & 72.29 & 37.03 & 5873
& \textbf{84.68} & 71.51 & 36.91 & 5839 \\
\cmidrule(lr){1-16}

\multirow{2}{*}{ProMoE}
& \multirow{2}{*}{Prev. layer} & Qwen3 & 96.0M
& \textbf{85.44} & 89.38 & \underline{64.38} & \underline{1792}
& \textbf{58.22} & 88.09 & \underline{64.21} & \underline{1779}
& \textbf{87.39} & 85.38 & \underline{60.71} & \underline{2003} \\
\cmidrule(lr){3-16}
& & GPT-OSS & 48.0M
& \underline{61.87} & \underline{93.54} & \textbf{67.94} & \textbf{2109}
& \textbf{43.74} & \underline{91.16} & \textbf{68.19} & \textbf{2032}
& \textbf{84.68} & \underline{92.95} & \underline{67.78} & \underline{2111} \\
\cmidrule(lr){1-16}

\multirow{2}{*}{FineMoE}
& \multirow{2}{*}{Prev. layer} & Qwen3 & --
& \textbf{85.44} & 76.67 & 54.84 & 2288
& \textbf{58.22} & 71.45 & 51.41 & 2447
& \textbf{87.39} & 74.08 & 51.40 & 2538 \\
\cmidrule(lr){3-16}
& & GPT-OSS & --
& \underline{61.87} & 85.63 & \underline{65.02} & \underline{2201}
& \textbf{43.74} & 71.96 & 50.40 & 3384
& \textbf{84.68} & 75.49 & 53.08 & 3188 \\
\cmidrule(lr){1-16}

\multirow{2}{*}{\makecell{Temporally-\\Extended MoE}}
& \multirow{2}{*}{Curr. layer} & Qwen3 & 52.0M
& 81.50 & \textbf{100.00} & 52.34 & 3299
& 51.04 & \textbf{100.00} & 49.58 & 3685
& \underline{84.36} & \textbf{100.00} & 58.77 & 2543 \\
\cmidrule(lr){3-16}
& & GPT-OSS & 23.0M
& 60.27 & \textbf{100.00} & 52.78 & 4275
& \underline{43.22} & \textbf{100.00} & 57.40 & 3546
& 64.29 & \textbf{100.00} & 62.61 & 2853 \\
\cmidrule(lr){1-16}

\multirow{2}{*}{\textbf{\makecell{Spatio-Temporal\\Router}}}
& \multirow{2}{*}{Prev. layer} & Qwen3 & 25.2M
& \underline{83.40} & \underline{90.62} & \textbf{69.03} & \textbf{1474}
& \underline{57.66} & \underline{88.40} & \textbf{65.36} & \textbf{1698}
& 84.11 & \underline{95.56} & \textbf{78.74} & \textbf{935} \\
\cmidrule(lr){3-16}
& & GPT-OSS & 4.4M
& \textbf{64.52} & 85.68 & 58.81 & 2951
& 42.46 & 89.12 & \underline{63.53} & \underline{2625}
& \underline{84.60} & 90.26 & \textbf{71.30} & \textbf{1736} \\

\bottomrule
\end{tabular}}
\caption{Main results on Qwen3 and GPT-OSS. LRU/LFU/LRFU use matched LM-only backbones ($sw=0$). Decision Availability denotes when all decision inputs become available: Prev. token permits a longer potential overlap window than Prev. layer; Curr. layer has no lookahead. Our full mode includes the last-to-first-layer token wrap-around. Added Params. counts new inference-time parameters, excluding external stores. Accuracy and hit rates are percentages; Load is MB/decode token. Results are five-seed means; bold/underline denote best/second-best values within each method category, backbone, task, and metric.}
\label{tab:main_results}
\end{table*}

\paragraph{Baselines and protocol.}
We compare LRU, LFU, and LRFU for cache update, and Least-Stale (SpecMD), ProMoE, FineMoE, and Temporally Extended MoE for prefetching (Section~\ref{sec:related_work}). Classical policies run on matched LM-only post-trained backbones using the same data, optimizer, and schedule; learned baselines follow their prescribed adaptation. Thus, we compare complete inference configurations, not policies on an identical routing trace. All methods share cache-accounting rules that charge every proactive insertion.

Main settings \((B,\tau,sw,R)\) are \((20,0.03,0.1,15)\) for Qwen3 and \((8,0.05,0.01,6)\) for GPT-OSS, where \(sw\) denotes the relevant cache-loss weight and \(R\) applies only to the full mode. These settings were fixed before task evaluation and shared across datasets. Tables~\ref{tab:main_results} and~\ref{tab:sw_ablation} report means over the same five distinct, fixed seeds. 

\paragraph{Training and decoding.}
Qwen3 and GPT-OSS use learning rates of \(10^{-5}\) and \(5\times10^{-6}\), respectively, for four post-training epochs. We use fused AdamW with cosine decay and 3\% warmup, training on the full training splits without validation-based checkpoint selection. Main runs update the backbone and auxiliary routers; LM-only controls use the same schedule without the cache loss or auxiliary routers. Training uses bfloat16, TF32, and gradient checkpointing, without LoRA or weight quantization. Experiments run on an eight-accelerator node with 140~GB of device memory per accelerator. We use weight decay 0.1, gradient clipping at 1.0, per-device batch size 1, and eight gradient-accumulation steps. Maximum sequence lengths are 512 for GSM8K and CommonsenseQA and 2048 for MATH. No cache-specific load-balancing loss is added. Evaluation uses each backbone's chat template and task-specific system instruction, with greedy decoding and the KV cache. Generation stops at EOS or a budget of 512, 1,024, or 64 new tokens for GSM8K, MATH, or CommonsenseQA, respectively. Additional evaluation and ablation details are in Appendix~\ref{app:implementation}.

\paragraph{Baseline settings.}
LRFU uses \(\lambda=0.5\). SpecMD/Least-Stale uses lookahead 3 and a prefetch budget equal to cache capacity. ProMoE uses training-split traces and lookahead 3. FineMoE uses a 1K-entry expert-map store built from training traces, lookahead 3, a cache-sized prefetch budget, 64 candidates, and retrieval top-\(k\) values of 8 for Qwen3 and 4 for GPT-OSS. Temporally Extended MoE uses a cache-sized option set, 128-dimensional controller and set embeddings, copy-router initialization, and its original losses.

\paragraph{Metrics.}
We report task accuracy, hard hit rate, load-adjusted hit rate, and expert-weight traffic. A trace-driven simulator synchronized with decoding counts routed expert accesses \(A\), demand misses \(D\), proactive loads \(P\), and decode tokens \(T\); shared experts are excluded. Each miss or insertion costs one complete expert transfer of size \(S_{\mathrm{exp}}\) MB:
\[
\mathrm{Hit}=1-\frac{D}{A},\qquad
\mathrm{AdjHit}=\frac{A-D}{A+P},\qquad
\mathrm{Load/token}=\frac{(D+P)S_{\mathrm{exp}}}{T}.
\]
Prefill initializes the decode cache, but its transfers are excluded. For update-only methods, \(P=0\) and AdjHit equals Hit, so it is omitted. Load/token is the primary algorithmic cost; AdjHit provides a normalized penalty for proactive traffic. These metrics measure decode-stage transfer demand, not end-to-end latency or transfer--computation overlap.

\subsection{Main Results}

\paragraph{Update-only retention.}
Table~\ref{tab:main_results} shows that Temporal Router achieves the highest hit rate and lowest Load among update-only configurations on all six backbone--task pairs. On Qwen3, gains over the strongest classical hit-rate baseline are 10.46, 10.70, and 33.34 points on GSM8K, MATH, and CommonsenseQA; Load falls from 1353 to 974, 1294 to 906, and 1512 to 304~MB/token. GPT-OSS shows the same ranking with 2.2M added inference-time parameters.

\paragraph{Prefetch-enabled refinement.}
Among evaluated prefetchers, Spatio-Temporal Router leads all Qwen3 tasks in adjusted hit rate and Load. Relative to ProMoE, it improves adjusted hit by 1.15--18.03 points and reduces traffic by 4.6--53.3\%. GPT-OSS results are mixed: the full mode leads CommonsenseQA and ranks second on MATH in both metrics; on GSM8K, it attains the highest accuracy but weaker traffic efficiency. The two operating modes are not interchangeable: Temporal Router has lower total Load in all six settings, while the full mode achieves higher pre-access coverage by spending proactive traffic.

\paragraph{Quality and overhead.}
The full mode adds 25.2M parameters on Qwen3 and 4.4M on GPT-OSS (0.083\% and 0.021\% inference-time model-size overhead), versus 96.0M and 48.0M for ProMoE. FineMoE adds no trainable predictor but uses an external expert-map store. Task quality must be read alongside traffic: on Qwen3, full-mode accuracies are 83.40\%, 57.66\%, and 84.11\%, compared with LM-only values of 85.44\%, 58.22\%, and 87.39\%. Thus, better residency does not imply unchanged task behavior. Raw hit alone is also insufficient: Temporally Extended MoE attains 100\% hit by restricting routing to its option set, yet has less favorable traffic and accuracy.

\section{Analysis and Discussion}
\label{sec:analysis}
\begin{table}[!t]
\centering
\scriptsize
\setlength{\tabcolsep}{2.1pt}
\renewcommand{\arraystretch}{1.08}

\resizebox{0.9\linewidth}{!}{%
\begin{tabular}{@{}cccc@{\hspace{6pt}}!{\vrule width 0.5pt}@{\hspace{6pt}}lcccc@{}}
\toprule
\multicolumn{4}{c@{\hspace{6pt}}!{\vrule width 0.5pt}@{\hspace{6pt}}}{\textbf{Temporal Router}} &
\multicolumn{5}{c}{\textbf{Spatio-Temporal Router}} \\
\cmidrule(r{5pt}){1-4}
\cmidrule(r{5pt}){5-9}
\textbf{Setting} &
\textbf{Acc.} &
\makecell{\textbf{Hit}} &
\makecell{\textbf{Load}} &
\textbf{Setting} &
\textbf{Acc.} &
\makecell{\textbf{Hit}} &
\makecell{\textbf{Adj. Hit}} &
\makecell{\textbf{Load}} \\
\midrule

MoE / LRU
& 85.44 & 61.19 & 1406.50
& MoE / LRU
& 85.44 & 61.19 & -- & 1406.50 \\
\cmidrule(r{5pt}){1-4}
\cmidrule(r{5pt}){5-9}

\makecell{\(sw=0.1\)\\auxiliary only}
& \AblCell{85.44}{\AblEq{0.00}}
& \AblCell{63.01}{\AblUpRed{1.82}}
& \AblCell{1340.71}{\AblDownRed{65.79}}
& \makecell{\(sw=0.1\)\\auxiliary only}
& \AblCell{85.44}{\AblEq{0.00}}
& \AblCell{67.44}{\AblUpRed{6.25}}
& \AblCell{62.53}{\AblUpRed{1.34}}
& \AblCell{1465.59}{\AblUpBlue{59.09}} \\
\cmidrule(r{5pt}){1-4}
\cmidrule(r{5pt}){5-9}

\makecell{\(sw=0.1\)\\temporal only}
& \AblCell{85.44}{\AblEq{0.00}}
& \AblCell{73.13}{\AblUpRed{11.94}}
& \AblCell{973.86}{\AblDownRed{432.64}}
& \makecell{\(sw=0.1\)\\spatio only}
& \AblCell{83.24}{\AblDownBlue{2.20}}
& \AblCell{93.04}{\AblUpRed{31.85}}
& \AblCell{66.94}{\AblUpRed{5.75}}
& \AblCell{1665.51}{\AblUpBlue{259.01}} \\
\cmidrule(r{5pt}){1-4}
\cmidrule(r{5pt}){5-9}

\(sw=0.1\)
& \AblCell{85.44}{\AblEq{0.00}}
& \AblCell{73.13}{\AblUpRed{11.94}}
& \AblCell{973.86}{\AblDownRed{432.64}}
& \(sw=0.1\)
& \AblCell{83.40}{\AblDownBlue{2.04}}
& \AblCell{90.62}{\AblUpRed{29.43}}
& \AblCell{69.03}{\AblUpRed{7.84}}
& \AblCell{1474.00}{\AblUpBlue{67.50}} \\

\(sw=0.3\)
& \AblCell{84.99}{\AblDownBlue{0.45}}
& \AblCell{79.02}{\AblUpRed{17.83}}
& \AblCell{760.30}{\AblDownRed{646.20}}
& \(sw=0.3\)
& \AblCell{83.47}{\AblDownBlue{1.97}}
& \AblCell{93.67}{\AblUpRed{32.48}}
& \AblCell{73.42}{\AblUpRed{12.23}}
& \AblCell{1229.04}{\AblDownRed{177.46}} \\

\(sw=0.5\)
& \AblCell{82.56}{\AblDownBlue{2.88}}
& \AblCell{81.98}{\AblUpRed{20.79}}
& \AblCell{653.07}{\AblDownRed{753.43}}
& \(sw=0.5\)
& \AblCell{81.80}{\AblDownBlue{3.64}}
& \AblCell{95.01}{\AblUpRed{33.82}}
& \AblCell{75.66}{\AblUpRed{14.47}}
& \AblCell{1107.47}{\AblDownRed{299.03}} \\

\(sw=1.0\)
& \AblCell{79.15}{\AblDownBlue{6.29}}
& \AblCell{85.31}{\AblUpRed{24.12}}
& \AblCell{532.33}{\AblDownRed{874.17}}
& \(sw=1.0\)
& \AblCell{81.35}{\AblDownBlue{4.09}}
& \AblCell{96.34}{\AblUpRed{35.15}}
& \AblCell{78.00}{\AblUpRed{16.81}}
& \AblCell{985.16}{\AblDownRed{421.34}} \\

\(sw=2.0\)
& \AblCell{74.60}{\AblDownBlue{10.84}}
& \AblCell{88.44}{\AblUpRed{27.25}}
& \AblCell{418.89}{\AblDownRed{987.61}}
& \(sw=2.0\)
& \AblCell{76.50}{\AblDownBlue{8.94}}
& \AblCell{97.41}{\AblUpRed{36.22}}
& \AblCell{79.67}{\AblUpRed{18.48}}
& \AblCell{900.91}{\AblDownRed{505.59}} \\

\bottomrule
\end{tabular}}
\caption{Qwen3/GSM8K ablations. MoE/LRU is the LM-only reference ($sw=0$); auxiliary-only freezes this backbone. Temporal-only repeats the standard Temporal Router; Spatio-only combines prefetching with LRU eviction. Other $sw$ rows jointly train the backbone and auxiliary routers. Accuracy/hit rates are percentages; Load is MB/token. Five-seed means are shown, with changes from MoE/LRU in parentheses. Red marks higher hit or lower traffic; blue marks lower accuracy or higher traffic.}
\label{tab:sw_ablation}
\end{table}

\paragraph{Adaptation scope.}
Table~\ref{tab:sw_ablation} compares auxiliary-only and joint adaptation. At \(sw=0.1\), auxiliary-only training preserves accuracy: Temporal Router reaches 63.01\% hit at 1340.71~MB/token; the full mode reaches 67.44\% hit and 62.53\% adjusted hit at 1465.59~MB/token. Joint training raises Temporal Router to 73.13\% hit at 973.86~MB/token without accuracy loss. The full mode reaches 90.62\% hit and 69.03\% adjusted hit at 1474.00~MB/token, with a 2.04-point accuracy drop. Relative to auxiliary-only, its gain is higher coverage, not lower traffic (+8.41~MB/token).

\paragraph{Router composition.}
Temporal-only repeats the standard Temporal Router at \(sw=0.1\). Spatio-only removes \(p^T\), trains with \(s^S_{t,l}=p^R_{t-1,l}+p^S_{\rho(t,l)}\), and retains cross-layer prefetching. Demand-loaded experts enter the cache with LRU eviction rather than learned temporal retention. Relative to the full mode, hard hit rises (93.04\% versus 90.62\%), but adjusted hit falls (66.94\% versus 69.03\%) and Load increases (1665.51 versus 1474.00~MB/token). The combined retention--refinement configuration thus uses less traffic than LRU-backed Spatio-only, despite lower immediate coverage.

\paragraph{Proactive traffic.}
Higher raw hit need not imply lower total Load, since every prefetch transfer is counted. With fixed traces and equal transfer costs, exposed loading ranges from demand-only to serialized demand-plus-prefetch cost. Hit and AdjHit measure coverage, not latency bounds; Appendix~\ref{app:overlap} gives the assumptions.

\begin{table*}[!t]
\centering
\scriptsize
\setlength{\tabcolsep}{4.2pt}
\renewcommand{\arraystretch}{1.13}
\resizebox{0.9\textwidth}{!}{%
\begin{tabular}{@{}c|ccc@{\hspace{7pt}}!{\vrule width 0.5pt}@{\hspace{7pt}}c|ccc@{}}
\toprule
\multirow{2}{*}{\textbf{\(B\)}} &
\multicolumn{3}{c@{\hspace{7pt}}!{\vrule width 0.5pt}@{\hspace{7pt}}}{\textbf{Temporal Router}} &
\multicolumn{4}{c}{\textbf{Spatio-Temporal Router}} \\
\cmidrule(lr){2-4}
\cmidrule(lr){5-8}
& \textbf{Acc.} & \textbf{Hit} & \textbf{Load} & \textbf{Acc.} &
\makecell{\textbf{Low \(R\)}\\Hit / Adj. / Load} &
\makecell{\textbf{Mid \(R\)}\\Hit / Adj. / Load} &
\makecell{\textbf{Full \(R=B\)}\\Hit / Adj. / Load} \\
\midrule
12 & 84.46 & 62.91 & 1344 & 85.44 &
\makecell{\(R=4\)\\66.53 / 63.56 / 1382} &
\makecell{\(R=8\)\\73.92 / 65.12 / 1435} &
\makecell{\(R=12\)\\80.79 / 60.63 / 1901} \\
20 & 85.44 & 73.13 & 974 & 83.40 &
\makecell{\(R=8\)\\81.84 / 74.51 / 1015} &
\makecell{\(R=15\)\\90.62 / 69.03 / 1474} &
\makecell{\(R=20\)\\93.49 / 57.93 / 2461} \\
30 & 85.37 & 80.00 & 725 & 83.85 &
\makecell{\(R=8\)\\86.14 / 80.40 / 761} &
\makecell{\(R=20\)\\95.11 / 71.14 / 1398} &
\makecell{\(R=30\)\\97.23 / 53.39 / 3077} \\
\bottomrule
\end{tabular}}
\caption{Cache capacity $B$ and refinement budget $R$ on Qwen3/GSM8K; the main setting is $B=20,R=15$. Spatio-Temporal cells list $R$ above Hit/Adj.~Hit/Load. Budgets at each $B$ replay one checkpoint with shared accuracy. Metrics are percentages except Load (MB/token).}
\label{tab:cache_budget_sweep}
\end{table*}

\paragraph{Cache capacity and refinement budget.}
Table~\ref{tab:cache_budget_sweep} studies \(B\) and \(R\) on Qwen3/GSM8K. For Temporal Router, increasing \(B\) from 12 to 30 raises hit from 62.91\% to 80.00\% and reduces Load from 1344 to 725~MB/token, with accuracy within one point. At common \(R=8\), the full mode's hard/adjusted hit rises from 73.92\%/65.12\% to 86.14\%/80.40\%, and Load falls from 1435 to 761~MB/token. Because each \(B\) uses a separate checkpoint, its effects include model adaptation.

Within each \(B\), replaying the same checkpoint isolates inference-time \(R\). At \(B=20\), raising \(R\) from 8 to 15 and 20 increases hit from 81.84\% to 90.62\% and 93.49\%, but decreases adjusted hit from 74.51\% to 69.03\% and 57.93\% and raises Load from 1015 to 1474 and 2461~MB/token. Across all three capacities, \(R=B\) yields the highest raw hit but worst traffic efficiency. The main \(R=15\) setting favors coverage, while smaller budgets offer lower-traffic operating points without retraining.

\paragraph{Cache-loss weight.}
Table~\ref{tab:sw_ablation} varies \(sw\) on Qwen3/GSM8K as a post hoc sensitivity study, not task-specific tuning of the main settings. For Temporal Router, increasing \(sw\) from 0.1 to 2.0 raises hit from 73.13\% to 88.44\% and lowers Load from 973.86 to 418.89~MB/token. Relative to MoE/LRU, these settings improve hit by 11.94--27.25 points and reduce traffic by 432.64--987.61~MB/token. The accuracy cost grows: \(sw=0.1\) preserves 85.44\% accuracy, \(sw=0.3\) loses 0.45 points, and \(sw=2.0\) loses 10.84 points.

For the full mode, stronger supervision raises hard hit from 90.62\% to 97.41\% and adjusted hit from 69.03\% to 79.67\%. Demand-traffic savings outweigh proactive insertions from \(sw=0.3\) onward: \(sw=0.1\) increases baseline Load by 67.50~MB/token, while \(sw=0.3\) and \(sw=2.0\) reduce it by 177.46 and 505.59~MB/token. Accuracy losses are 1.97--2.04 points at the two smaller weights and 8.94 points at \(sw=2.0\). Moderate supervision therefore offers a useful quality--traffic trade-off, whereas large weights over-optimize cache locality at the expense of task quality.

\begin{figure*}[!t]
\centering
\captionsetup[subfigure]{font=small,labelfont=bf}
\setlength{\tabcolsep}{2pt}

\begin{subfigure}[t]{0.31\textwidth}
    \centering
    \includegraphics[width=\linewidth]{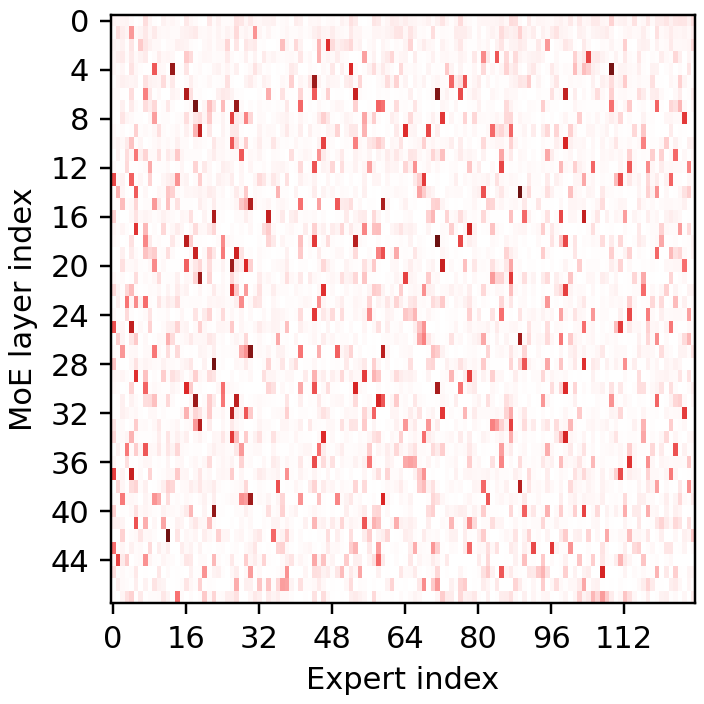}
    \caption{Temporal, \(sw=0.1\)}
    \label{fig:heatmap_temporal_sw01}
\end{subfigure}
\hfill
\begin{subfigure}[t]{0.31\textwidth}
    \centering
    \includegraphics[width=\linewidth]{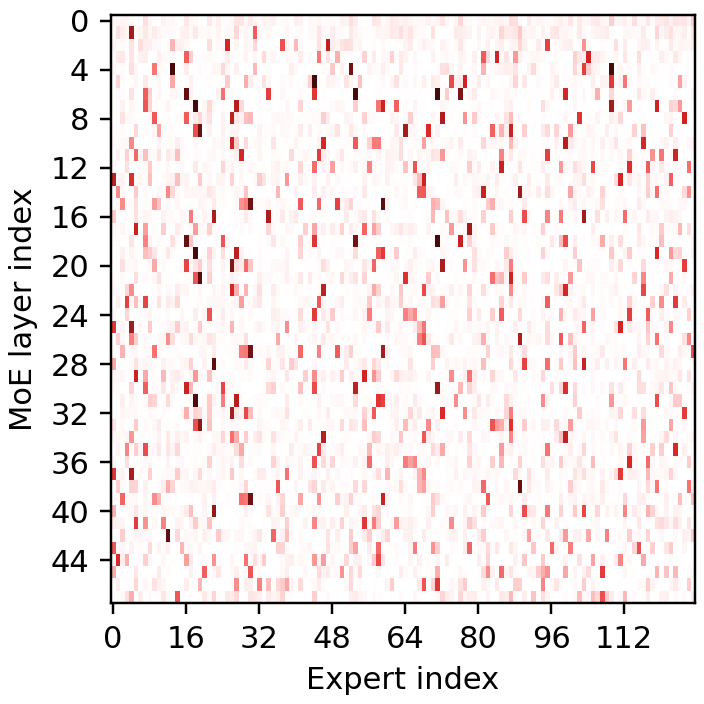}
    \caption{Temporal, \(sw=0.5\)}
    \label{fig:heatmap_temporal_sw05}
\end{subfigure}
\hfill
\begin{subfigure}[t]{0.31\textwidth}
    \centering
    \includegraphics[width=\linewidth]{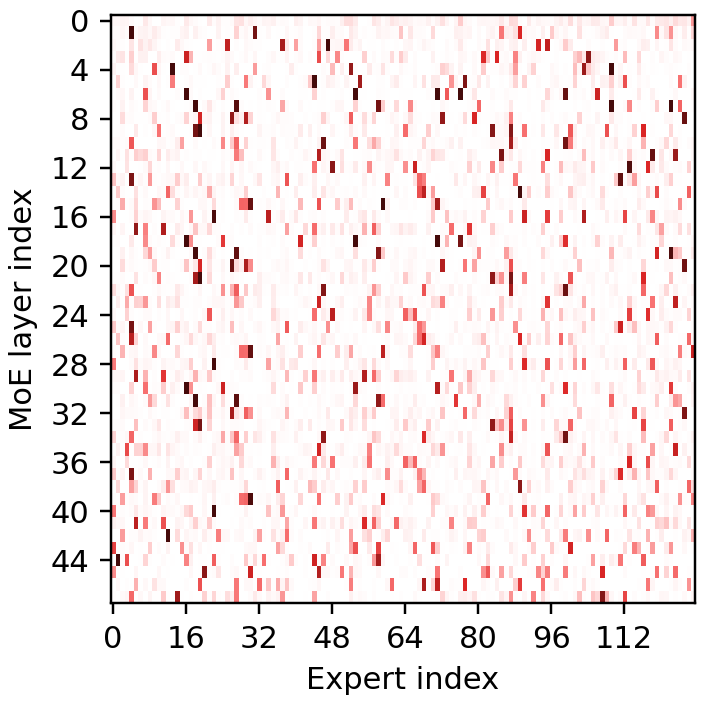}
    \caption{Temporal, \(sw=2.0\)}
    \label{fig:heatmap_temporal_sw20}
\end{subfigure}

\vspace{3pt}

\begin{subfigure}[t]{0.31\textwidth}
    \centering
    \includegraphics[width=\linewidth]{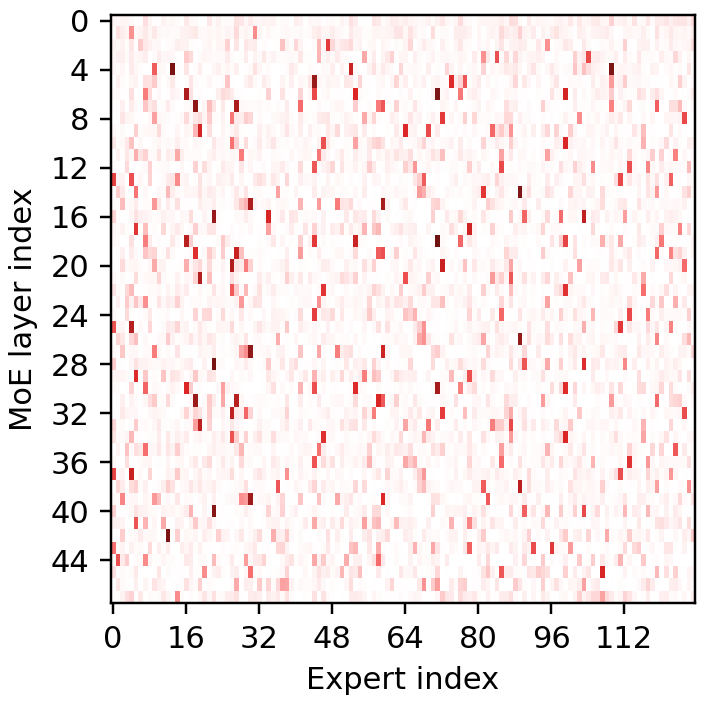}
    \caption{Spatio-Temporal, \(sw=0.1\)}
    \label{fig:heatmap_spatiotemporal_sw01}
\end{subfigure}
\hfill
\begin{subfigure}[t]{0.31\textwidth}
    \centering
    \includegraphics[width=\linewidth]{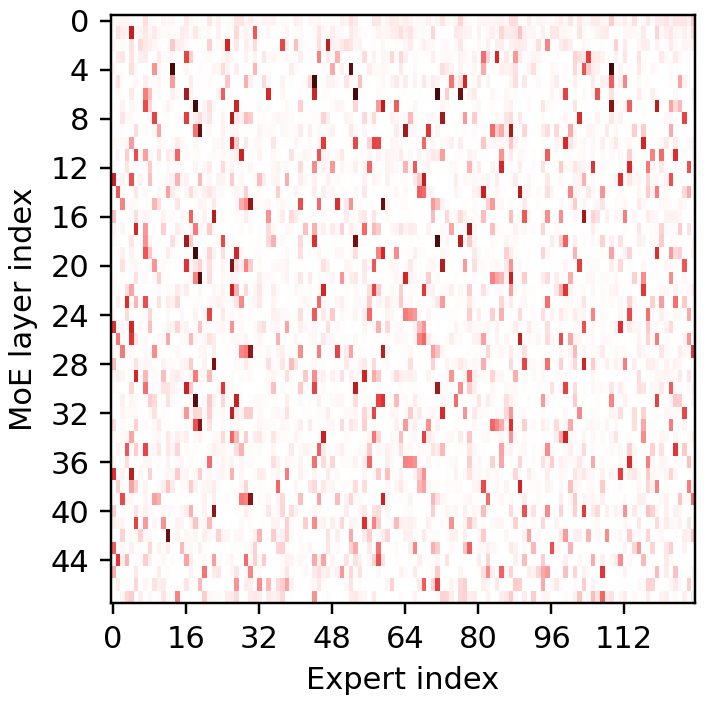}
    \caption{Spatio-Temporal, \(sw=0.5\)}
    \label{fig:heatmap_spatiotemporal_sw05}
\end{subfigure}
\hfill
\begin{subfigure}[t]{0.31\textwidth}
    \centering
    \includegraphics[width=\linewidth]{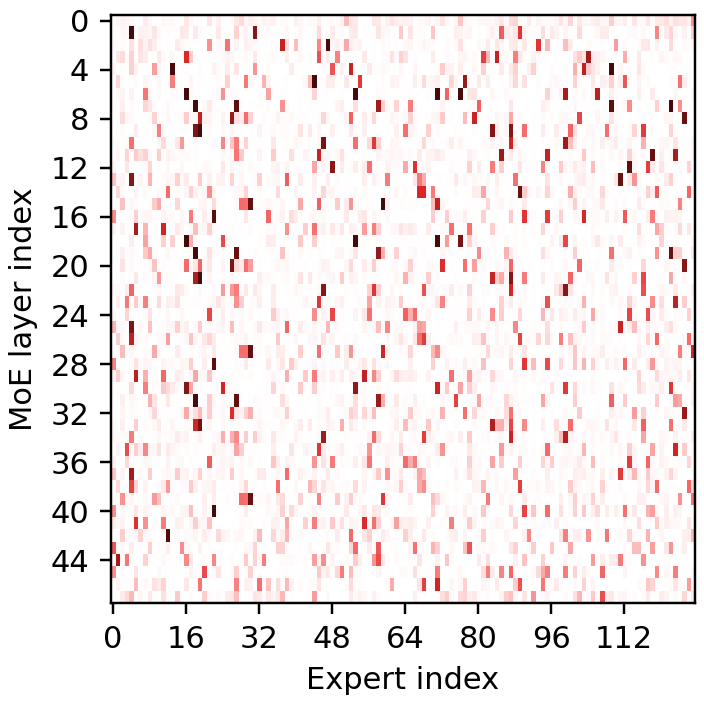}
    \caption{Spatio-Temporal, \(sw=2.0\)}
    \label{fig:heatmap_spatiotemporal_sw20}
\end{subfigure}

\caption{Native expert-selection frequency on Qwen3/GSM8K. Rows show Temporal Router (top) and Spatio-Temporal Router (bottom); darker cells indicate more frequent selection.}
\label{fig:expert_usage_heatmaps}
\end{figure*}

\begin{figure*}[!t]
\centering
\setlength{\tabcolsep}{1pt}

\begin{subfigure}[t]{0.245\textwidth}
    \centering
    \includegraphics[width=\linewidth]{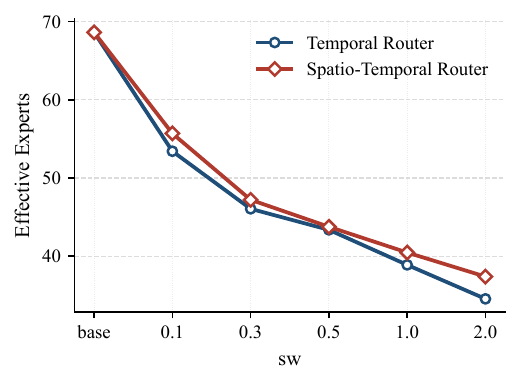}
    \caption{Effective experts}
    \label{fig:sw_effective_experts}
\end{subfigure}
\hfill
\begin{subfigure}[t]{0.245\textwidth}
    \centering
    \includegraphics[width=\linewidth]{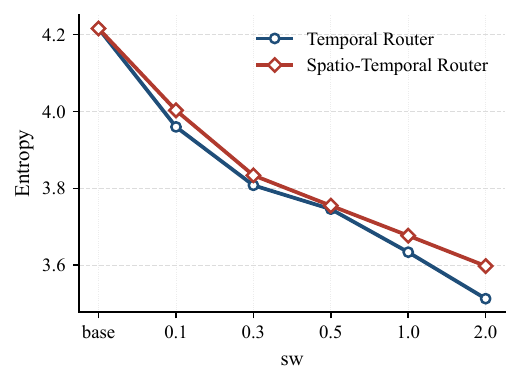}
    \caption{Entropy}
    \label{fig:sw_entropy}
\end{subfigure}
\hfill
\begin{subfigure}[t]{0.245\textwidth}
    \centering
    \includegraphics[width=\linewidth]{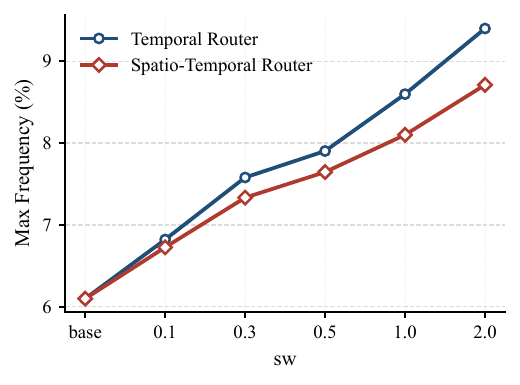}
    \caption{Max frequency}
    \label{fig:sw_max_frequency}
\end{subfigure}
\hfill
\begin{subfigure}[t]{0.245\textwidth}
    \centering
    \includegraphics[width=\linewidth]{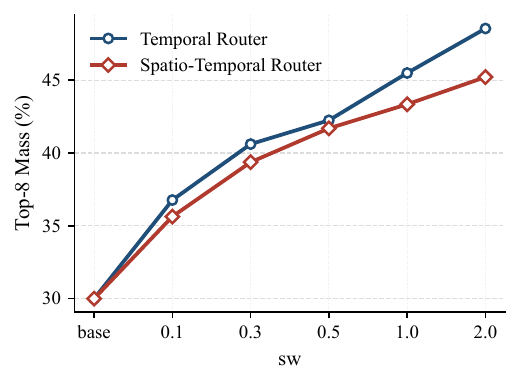}
    \caption{Top-8 mass}
    \label{fig:sw_top8_mass}
\end{subfigure}

\caption{Expert-use concentration versus cache-loss weight, averaged over MoE layers. Base is the LM-only reference. Per-layer effective-expert count is $\exp(\mathrm{entropy})$; max frequency and top-8 mass measure the most-used expert's frequency and the eight most-used experts' total frequency.}
\label{fig:sw_routing_behavior}
\end{figure*}

\paragraph{Expert-use concentration.}
Figure~\ref{fig:expert_usage_heatmaps} shows increasingly concentrated native expert use as \(sw\) grows. Auxiliary routers do not override the native Top-\(K\) rule, but joint post-training can reshape the distribution on which that rule operates. The observed concentration is consistent with improved cache locality, although it does not by itself establish greater temporal stability.

Figure~\ref{fig:sw_routing_behavior} quantifies this trend. For Temporal Router, increasing \(sw\) from 0.1 to 2.0 raises maximum expert frequency from 6.82\% to 9.40\% and top-8 mass from 36.76\% to 48.56\%, while entropy falls from 3.96 to 3.51 and effective experts from 53.41 to 34.54. The full mode changes in the same direction with flatter curves: maximum frequency rises from 6.73\% to 8.71\%, and top-8 mass from 35.63\% to 45.22\%. Together with the auxiliary-only control, these observations support joint adaptation toward more cache-compatible demand. They characterize aggregate model behavior, not individual-router contributions. Moderate supervision retains substantial routing diversity; excessive concentration accompanies the accuracy losses at large \(sw\).

\section{Related Work}
\label{sec:related_work}
\paragraph{LLM and MoE serving.}
FlexGen~\citep{sheng2023flexgen} and vLLM/PagedAttention~\citep{kwon2023efficient} address memory pressure through tensor placement, KV-cache management, and batching. MoE serving additionally manages large, sparse, dynamically selected expert weights. DeepSpeed-MoE~\citep{rajbhandari2022deepspeed} and Tutel~\citep{hwang2023tutel} optimize parallel execution, while MoE-Infinity~\citep{xue2024moe} targets offloading through tracing, caching, and prefetching. Related work studies serverless placement and speculative scheduling~\citep{yu2026moeless,li2025speculative}. Unlike these serving systems, we learn model-side cache priorities.

\paragraph{Expert caching and prefetching.}
LRU and LFU use recency and frequency~\citep{mattson1970evaluation,robinson1990data}; LRFU interpolates between them~\citep{lee2001lrfu}. MoE-specific methods exploit richer signals: Least-Stale uses access history~\citep{hoang2026specmd}, ProMoE trains an activation-based predictor~\citep{song2024promoe}, FineMoE retrieves expert maps~\citep{yu2025finemoe}, and Temporally Extended MoE learns whether to retain or switch expert sets~\citep{shen2026temporally}. We instead learn cache priorities jointly with model adaptation.

\paragraph{Spatio-temporal prefetching.}
STEP uses adaptive spatio-temporal expert prefetching~\citep{liu2026step}; ST-MoE combines profiled cross-layer and consecutive-token correlations with table-based prediction and reconfigurable hardware~\citep{zhao2026stmoe}. Both emphasize inference-time prediction or system support. We focus on cache-aware model adaptation and support both update-only retention and bounded pre-access refinement. Their reported latency reflects integrated runtime or hardware pipelines, so we compare designs qualitatively rather than treating those measurements as directly comparable to our traffic metric.

\Needspace{10\baselineskip}
\section{Conclusion}
We presented cache-aware post-training for expert-cache management in memory-constrained MoE inference. Temporal Router learns same-layer retention without proactive loading; Spatio-Temporal Router adds bounded causal refinement while preserving the native inference-time Top-\(K\) rule. Across two backbones and three tasks, Temporal Router consistently reduces transfer demand relative to matched LM-only replacement configurations. The full mode offers stronger pre-access coverage, with favorable traffic results on Qwen3 and task-dependent results on GPT-OSS. Auxiliary-only and component ablations support the value of joint model adaptation and temporal retention, while the budget sweeps distinguish the roles of cache capacity, refinement aggressiveness, and cache-loss weight. These controls expose a quality--coverage--traffic trade-off rather than a single universally optimal configuration. The results characterize algorithmic transfer demand; end-to-end performance remains dependent on the runtime, as discussed in Appendix~\ref{app:limitations}.

\FloatBarrier



\bibliography{iclr2027_conference}

@article{xue2024moe,
  title={Moe-infinity: Offloading-efficient moe model serving},
  author={Xue, Leyang and Fu, Yao and Lu, Zhan and Mai, Luo and Marina, Mahesh},
  journal={arXiv e-prints},
  pages={arXiv--2401},
  year={2024}
}

@article{song2024promoe,
  title={Promoe: Fast moe-based llm serving using proactive caching},
  author={Song, Xiaoniu and Zhong, Zihang and Chen, Rong and Chen, Haibo},
  journal={arXiv preprint arXiv:2410.22134},
  year={2024}
}

@article{lee2001lrfu,
  title={LRFU: A spectrum of policies that subsumes the least recently used and least frequently used policies},
  author={Lee, Donghee and Choi, Jongmoo and Kim, Jong-Hun and Noh, Sam H and Min, Sang Lyul and Cho, Yookun and Kim, Chong Sang},
  journal={IEEE transactions on Computers},
  volume={50},
  number={12},
  pages={1352--1361},
  year={2001},
  publisher={IEEE Computer Society}
}

@article{hoang2026specmd,
  title={SpecMD: A Comprehensive Study On Speculative Expert Prefetching},
  author={Hoang, Duc and Jaiswal, Ajay and Samragh, Mohammad and Cho, Minsik},
  journal={arXiv preprint arXiv:2602.03921},
  year={2026}
}

@article{shen2026temporally,
  title={Temporally Extended Mixture-of-Experts Models},
  author={Shen, Zeyu and Henderson, Peter},
  journal={arXiv preprint arXiv:2604.20156},
  year={2026}
}

@article{cobbe2021training,
  title={Training verifiers to solve math word problems},
  author={Cobbe, Karl and Kosaraju, Vineet and Bavarian, Mohammad and Chen, Mark and Jun, Heewoo and Kaiser, Lukasz and Plappert, Matthias and Tworek, Jerry and Hilton, Jacob and Nakano, Reiichiro and others},
  journal={arXiv preprint arXiv:2110.14168},
  year={2021}
}

@article{hendrycks2021measuring,
  title={Measuring mathematical problem solving with the math dataset},
  author={Hendrycks, Dan and Burns, Collin and Kadavath, Saurav and Arora, Akul and Basart, Steven and Tang, Eric and Song, Dawn and Steinhardt, Jacob},
  journal={arXiv preprint arXiv:2103.03874},
  year={2021}
}

@inproceedings{talmor2019commonsenseqa,
  title={Commonsenseqa: A question answering challenge targeting commonsense knowledge},
  author={Talmor, Alon and Herzig, Jonathan and Lourie, Nicholas and Berant, Jonathan},
  booktitle={Proceedings of the 2019 Conference of the North American Chapter of the Association for Computational Linguistics: Human Language Technologies, Volume 1 (Long and Short Papers)},
  pages={4149--4158},
  year={2019}
}

@misc{qwen3technicalreport,
      title={Qwen3 Technical Report}, 
      author={Qwen Team},
      year={2025},
      eprint={2505.09388},
      archivePrefix={arXiv},
      primaryClass={cs.CL},
      url={https://arxiv.org/abs/2505.09388}, 
}

@article{agarwal2025gpt,
  title={gpt-oss-120b \& gpt-oss-20b model card},
  author={Agarwal, Sandhini and Ahmad, Lama and Ai, Jason and Altman, Sam and Applebaum, Andy and Arbus, Edwin and Arora, Rahul K and Bai, Yu and Baker, Bowen and Bao, Haiming and others},
  journal={arXiv preprint arXiv:2508.10925},
  year={2025}
}

@article{mattson1970evaluation,
  title={Evaluation techniques for storage hierarchies},
  author={Mattson, Richard L. and Gecsei, Jan and Slutz, Donald R. and Traiger, Irving L.},
  journal={IBM Systems journal},
  volume={9},
  number={2},
  pages={78--117},
  year={1970},
  publisher={IBM}
}

@inproceedings{robinson1990data,
  title={Data cache management using frequency-based replacement},
  author={Robinson, John T and Devarakonda, Murthy V},
  booktitle={Proceedings of the 1990 ACM SIGMETRICS conference on Measurement and modeling of computer systems},
  pages={134--142},
  year={1990}
}

@inproceedings{yu2025finemoe,
  author       = {Hanfei Yu and
                  Xingqi Cui and
                  Hong Zhang and
                  Hao Wang and
                  Hao Wang},
  editor       = {Antonio Barbalace and
                  Luo Mai and
                  Roxana Geambasu and
                  Peter R. Pietzuch},
  title        = {Taming Latency-Memory Trade-Off in MoE-Based {LLM} Serving via Fine-Grained
                  Expert Offloading},
  booktitle    = {Proceedings of the 21st European Conference on Computer Systems, EuroSys
                  2026, McEwan Hall/The University of Edinburgh, Edinburgh, Scotland,
                  UK, April 27-30, 2026},
  pages        = {176--191},
  publisher    = {{ACM}},
  year         = {2026},
}

@inproceedings{sheng2023flexgen,
  title={Flexgen: High-throughput generative inference of large language models with a single gpu},
  author={Sheng, Ying and Zheng, Lianmin and Yuan, Binhang and Li, Zhuohan and Ryabinin, Max and Chen, Beidi and Liang, Percy and R{\'e}, Christopher and Stoica, Ion and Zhang, Ce},
  booktitle={International Conference on Machine Learning},
  pages={31094--31116},
  year={2023},
  organization={PMLR}
}

@inproceedings{kwon2023efficient,
  title={Efficient memory management for large language model serving with pagedattention},
  author={Kwon, Woosuk and Li, Zhuohan and Zhuang, Siyuan and Sheng, Ying and Zheng, Lianmin and Yu, Cody Hao and Gonzalez, Joseph and Zhang, Hao and Stoica, Ion},
  booktitle={Proceedings of the 29th symposium on operating systems principles},
  pages={611--626},
  year={2023}
}

@inproceedings{rajbhandari2022deepspeed,
  title={Deepspeed-moe: Advancing mixture-of-experts inference and training to power next-generation ai scale},
  author={Rajbhandari, Samyam and Li, Conglong and Yao, Zhewei and Zhang, Minjia and Aminabadi, Reza Yazdani and Awan, Ammar Ahmad and Rasley, Jeff and He, Yuxiong},
  booktitle={International conference on machine learning},
  pages={18332--18346},
  year={2022},
  organization={PMLR}
}

@article{hwang2023tutel,
  title={Tutel: Adaptive mixture-of-experts at scale},
  author={Hwang, Changho and Cui, Wei and Xiong, Yifan and Yang, Ziyue and Liu, Ze and Hu, Han and Wang, Zilong and Salas, Rafael and Jose, Jithin and Ram, Prabhat and others},
  journal={Proceedings of Machine Learning and Systems},
  volume={5},
  pages={269--287},
  year={2023}
}

@article{yu2026moeless,
  title={MoEless: Efficient MoE LLM Serving via Serverless Computing},
  author={Yu, Hanfei and Ouyang, Bei and He, Shwai and Li, Ang and Wang, Hao},
  journal={arXiv preprint arXiv:2603.06350},
  year={2026}
}

@article{li2025speculative,
  title={Speculative MoE: Communication efficient parallel moe inference with speculative token and expert pre-scheduling},
  author={Li, Yan and Zheng, Pengfei and Chen, Shuang and Xu, Zewei and Lai, Yuanhao and Du, Yunfei and Wang, Zhengang},
  journal={arXiv preprint arXiv:2503.04398},
  year={2025}
}

@misc{zhao2026stmoe,
      title={A Spatio-Temporal Expert Prefetching Framework for Efficient MoE-based LLM Inference}, 
      author={Yingnan Zhao and Razvan Bunescu and Ahmed Louri and Avinash Karanth and Ke Wang},
      year={2026},
      eprint={2606.15453} 
}

@INPROCEEDINGS{liu2026step,
  author={Liu, Fangxin and Yang, Ning and Wang, Zongwu and Guan, Chenyang and Li, Haomin and Feng, Yu and Lu, Liqiang and Li, Xiang and Yang, Siran and Wang, Jiamang and Qu, Lin and Jiang, Li and Guan, Haibing},
  booktitle={2026 ACM/IEEE 53rd Annual International Symposium on Computer Architecture (ISCA)}, 
  title={STEP: Adaptive Spatio-Temporal Expert Prefetching for Low-Latency and Memory-Efficient MoE Inference}, 
  year={2026},
  volume={},
  number={},
  pages={1336-1350}}
\bibliographystyle{iclr2027_conference}

\clearpage
\appendix

\section{Full Inference Algorithms}
\label{app:algorithms}
Algorithm~\ref{alg:spatio-temporal-router} gives the full prefetch-enabled procedure and invokes the shared retention operator in Algorithm~\ref{alg:update-cache}. Temporal Router admits only experts already resident or demand-loaded for the current computation; the full mode additionally charges each proactive insertion before target access.

\begin{algorithm}[H]
\caption{Full Spatio-Temporal Router Inference at \((t,l)\)}
\label{alg:spatio-temporal-router}
\begin{algorithmic}[1]
\Require $\mathcal{C}^{T}_{t,l}$; $p^R_{t-1,l},p^T_{t-1,l},p^S_{\rho(t,l)}$; budgets $R,B$
\State $a^T_{t,l}\gets p^R_{t-1,l}+p^T_{t-1,l}$
\State $s^{ST}_{t,l}\gets a^T_{t,l}+p^S_{\rho(t,l)}$
\State $\mathcal{A}\gets\operatorname{Top}_{R}(s^{ST}_{t,l})$
\State $\widetilde{\mathcal{C}}^{ST}_{t,l}\gets\mathcal{C}^{T}_{t,l}$
\ForAll{$e\in\mathcal{A}$ in descending order of $s^{ST}_{t,l,e}$}
    \If{$e\notin\widetilde{\mathcal{C}}^{ST}_{t,l}$}
        \If{$|\widetilde{\mathcal{C}}^{ST}_{t,l}|<B$}
            \State $\widetilde{\mathcal{C}}^{ST}_{t,l}\gets\widetilde{\mathcal{C}}^{ST}_{t,l}\cup\{e\}$ \Comment{proactive load}
        \Else
            \State $v\gets\arg\min_{u\in\widetilde{\mathcal{C}}^{ST}_{t,l}}s^{ST}_{t,l,u}$
            \If{$s^{ST}_{t,l,e}>s^{ST}_{t,l,v}$}
                \State $\widetilde{\mathcal{C}}^{ST}_{t,l}\gets(\widetilde{\mathcal{C}}^{ST}_{t,l}\setminus\{v\})\cup\{e\}$ \Comment{proactive load}
            \EndIf
        \EndIf
    \EndIf
\EndFor
\State Compute $p^R_{t,l}$, $S_{t,l}=\operatorname{TopK}(p^R_{t,l})$, and $p^T_{t,l}$ from $h_{t,l}$
\State Demand-load $S_{t,l}\setminus\widetilde{\mathcal{C}}^{ST}_{t,l}$ and execute the MoE layer
\State $s^T_{t,l}\gets p^R_{t,l}+p^T_{t,l}$
\State $\mathcal{C}^{T}_{t+1,l}\gets\Call{UpdateCache}{\widetilde{\mathcal{C}}^{ST}_{t,l},S_{t,l},s^T_{t,l},B}$
\State \Return $\mathcal{C}^{T}_{t+1,l}$
\end{algorithmic}
\end{algorithm}

\FloatBarrier

\section{Implementation and Evaluation Details}
\label{app:implementation}
\subsection{Initialization and evaluation controls}
The copy-initialization rules in Section~\ref{sec:training_deployment} apply before cache-aware training. For Tables~\ref{tab:main_results} and~\ref{tab:sw_ablation}, every configuration runs once under each of the same five distinct, fixed seeds. Metrics are computed per run and then averaged across the five runs. The main refinement budgets are \(R=15\) for Qwen3 and \(R=6\) for GPT-OSS, both 75\% of cache capacity. As described in the main text, \(R\) bounds the candidate set rather than the number of actual loads. The post-prefill cache initializes decoding, while all reported hit-rate and traffic statistics exclude prefill transfers. Each demand miss and proactive insertion is charged as one complete expert-weight transfer, without modeling scheduling or overlap.

\subsection{Ablation and sensitivity configurations}
Auxiliary-only training freezes the LM-only backbone, including \(W^R\), and trains only \(W^T\), or \(W^T\) and \(W^S\). Temporal-only is the standard update-only mode, repeated in Table~\ref{tab:sw_ablation} rather than trained separately. Spatio-only removes Temporal Router and its learned cross-token retention stage. Its cache-coverage objective uses \(s^S_{t,l}=p^R_{t-1,l}+p^S_{\rho(t,l)}\), and causal cross-layer prefetching continues with the same candidate-selection and score-based pre-access refinement rule. Any selected expert absent after prefetching is demand-loaded and inserted into the cache, with LRU evictions to maintain capacity. In the full mode, the corresponding demand-miss replacement uses Temporal Router priorities instead. Spatio-only therefore removes learned temporal retention, not demand-miss insertion or all cache replacement.

Table~\ref{tab:cache_budget_sweep} uses separately trained Qwen3/GSM8K checkpoints for \(B\in\{12,20,30\}\). Each Spatio-Temporal checkpoint is replayed with its three listed inference-time budgets without retraining, so accuracy is shared within each \(B\). The parameter \(R\) corresponds to \texttt{replace\_lowest}; it is distinct from the native Top-\(K\) expert count. Both budget and loss-weight sweeps are post hoc sensitivity studies, not task-specific selection of the main hyperparameters.

\begin{figure}[!ht]
\centering
\includegraphics[width=\linewidth]{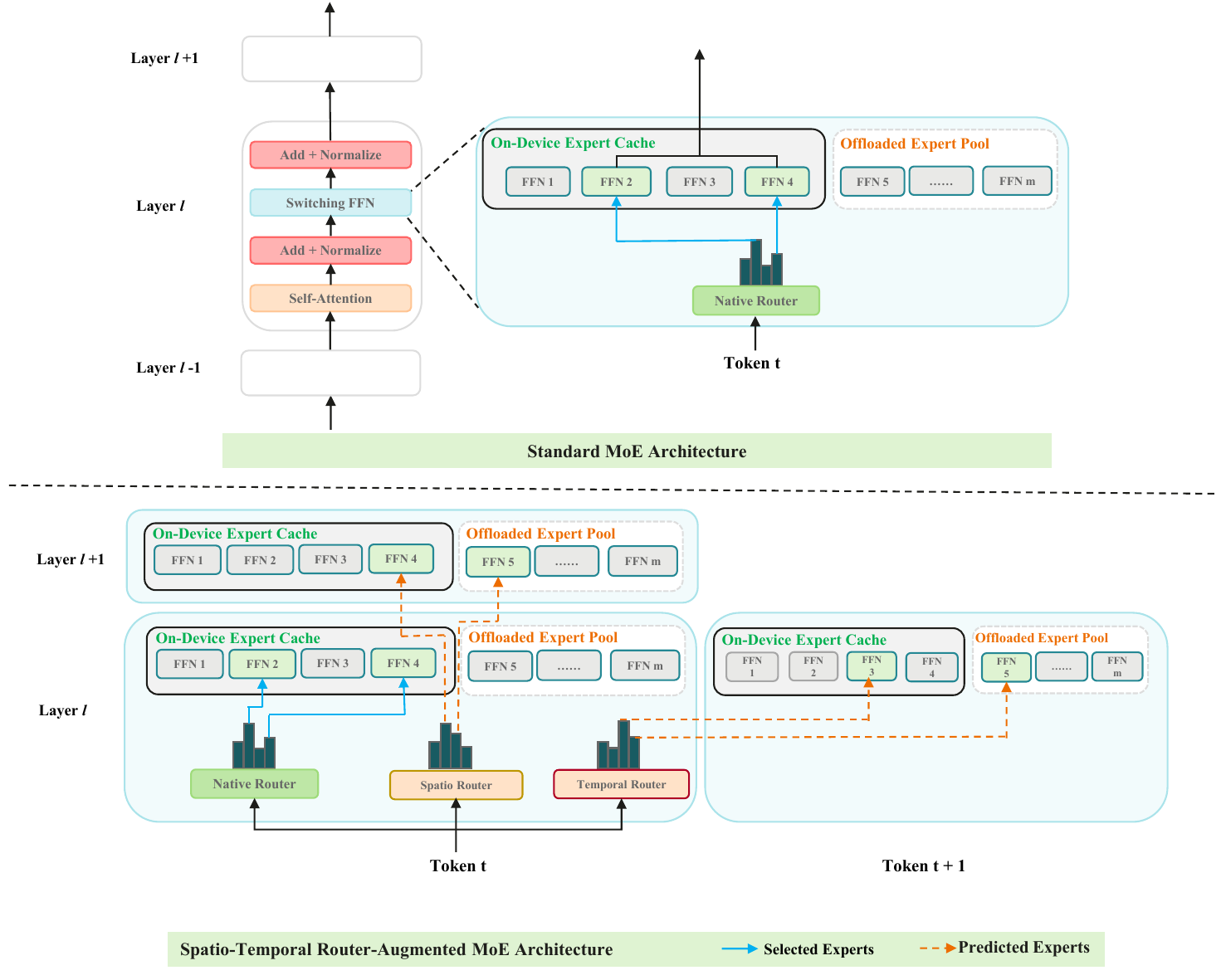}
\caption{Standard MoE (top) and Spatio-Temporal Router-augmented MoE (bottom), with on-device expert caches and offloaded pools. Solid blue arrows denote native selection; dashed orange arrows denote predicted demand, not transfers. Temporal Router retains only resident or demand-loaded experts, even when predictions include offloaded experts. Proactive loading occurs only during bounded pre-access refinement; expert indices and cache contents are illustrative.}
\label{fig:architecture_comparison}
\end{figure}
\FloatBarrier

\subsection{Interpreting asynchronous transfer overlap}
\label{app:overlap}
Total transferred bytes do not depend on whether transfers overlap computation. To illustrate how overlap can affect exposed loading cost, fix the routing and cache trace, including \(A,D,P,T\), and assume each expert transfer takes \(c_{\mathrm{exp}}\) seconds. Let \(\eta\in[0,1]\) denote the fraction of proactive transfer time hidden by useful computation. Treating demand transfers as fully exposed and ignoring transfer contention and auxiliary overhead, the modeled expert-loading cost per decode token is
\[
 t_{\mathrm{load}}(\eta)
 =\frac{c_{\mathrm{exp}}}{T}\bigl[D+(1-\eta)P\bigr].
\]
Complete prefetch overlap (\(\eta=1\)) leaves only demand-transfer cost; no overlap (\(\eta=0\)) charges both demand and proactive transfers. Accordingly,
\[
 \frac{D c_{\mathrm{exp}}}{T}
 \leq t_{\mathrm{load}}(\eta)
 \leq \frac{(D+P)c_{\mathrm{exp}}}{T}.
\]
These two endpoints can be written in terms of the reported metric definitions:
\begin{align*}
 t_{\mathrm{load}}(1)
 &=\frac{A c_{\mathrm{exp}}}{T}(1-\mathrm{Hit}),\\
 t_{\mathrm{load}}(0)
 &=\frac{(A+P)c_{\mathrm{exp}}}{T}(1-\mathrm{AdjHit}).
\end{align*}
The normalizers differ: Hit discounts demand misses over \(A\) accesses, whereas AdjHit charges all transfers over \(A+P\). Their numerical values therefore cannot be interpreted directly as the endpoints of a latency or speedup interval. Under the stated assumptions, a prefetch-enabled configuration can have higher total traffic yet lower exposed loading cost when enough proactive transfer time is hidden. The realized overlap fraction is not measured here, and perfect overlap may be infeasible because of limited lookahead or bandwidth. These bounds apply only to the simplified fixed-trace loading model, not to end-to-end latency; scheduling, contention, late prefetches, and auxiliary computation require runtime evaluation.

\FloatBarrier
\section{Detailed Architectural Comparison}
\label{app:architecture}
Figure~\ref{fig:architecture_comparison} compares the standard MoE block with its router-augmented counterpart. The native, Temporal, and Spatio Routers share the layer input hidden state. Native routing selects experts for the current computation; Temporal Router scores retention for the next token at the same layer, while Spatio Router predicts demand at the next MoE position. The update-only mode retains Temporal Router and omits Spatio Router.

\section{Limitations}
\label{app:limitations}
Our evaluation isolates algorithmic cache management rather than an end-to-end serving stack. Load/token measures simulated decode-stage expert-weight traffic; realized latency and throughput also depend on scheduling, transfer--computation overlap, interconnect bandwidth, and batching. Prefill initializes the cache but its transfers are excluded, so results characterize decode-stage rather than request-level efficiency. They establish neither wall-clock speedup nor energy savings.

Main runs require full-model post-training despite small inference-time parameter overhead. The auxiliary-only control supports joint adaptation but does not isolate changes in native-router parameters from other backbone changes. The full benchmark matrix uses one main cache capacity per backbone; capacity and refinement sensitivity are studied on Qwen3/GSM8K. Main comparisons and ablations use five seeds; the budget study replays fixed checkpoints.

Experiments focus on academic reasoning benchmarks and pure MoE models. Broader workloads, memory budgets, pretraining settings, and hybrid dense--sparse architectures remain untested. More concentrated native routing may improve locality but impair expert-parallel load balance. Finally, baseline algorithms are reproduced under common cache accounting rather than their original runtime environments; rankings may change under other workloads, memory hierarchies, or schedulers.

\end{document}